\documentclass[10pt, a4paper]{article}
\usepackage{graphicx}
\usepackage{verbatim}

\usepackage{lrec2026} 

\title{Who Benchmarks the Benchmarks? A Case Study of LLM Evaluation in Icelandic}

\name{\parbox{\textwidth}{\centering{Finnur Ágúst Ingimundarson$^{1}$, Steinunn Rut Friðriksdóttir$^{2}$, 
Bjarki Ármannsson$^{2,3}$, Iris Edda Nowenstein$^{2}$, Steinþór Steingrímsson$^{3}$}}}

\address{$^{1}$University of Zürich, $^{2}$University of Iceland, $^{3}$The Árni Magnússon Institute for Icelandic Studies  \\
 finnuragust.ingimundarson@uzh.ch, \{srf, irisen\}@hi.is, \{bjarki.armannsson,\\ steinthor.steingrimsson\}@arnastofnun.is}

\abstract{
This paper evaluates current Large Language Model (LLM) benchmarking for Icelandic, identifies problems, and calls for improved evaluation methods in low/medium-resource languages in particular. We show that benchmarks that include synthetic or machine-translated data that have not been verified in any way, commonly contain severely flawed test examples that are likely to skew the results and undermine the tests' validity. We warn against the use of such methods without verification in low/medium-resource settings as the translation quality can, at best, only be as good as MT quality for a given language at any given time. Indeed, the results of our quantitative error analysis on existing benchmarks for Icelandic show clear differences between human-authored/-translated benchmarks vs. synthetic or machine-translated benchmarks.
 \\ \newline \Keywords{benchmarking, machine translation, low-resource languages, Icelandic} }

\begin{document}

\hyphenation{development robust natural languages common-place EuroEval authors}

\maketitleabstract

\section{Introduction}
\label{sec:intro}
Benchmarking has been an integral part of development in NLP as a way of assessing performance on various tasks and traditionally on tasks with a single-label output, such as POS-tagging, NER, sentiment analysis, etc. With the advent of LLMs, that are not only capable of performing all of these traditional tasks and yet more \citep{brown2020gpt} but perhaps even of creating such data, the need for robust benchmarks is even greater than before. In their wake also follows a demand for creating new benchmarks that are better adapted to the capabilities of LLMs, i.e. beyond evaluating single-label output, and importantly, beyond tasks which purely target the scope of expertise of NLP researchers \citep{bowman-dahl-2021-will}.

This means that benchmark set creators increasingly find themselves creating benchmarks for natural language production and understanding in languages they do not speak and within subjects of which they are not experts. For example, there is a clear difference between constructing a NER evaluation metric in a language the researcher knows and generating a machine-translated benchmarking set for the evaluation of an LLMs' capabilities in medical text summarization \citep{alaa2025}. Furthermore, the field -- and therefore benchmarking -- has largely been dominated by English (and a few other high-resource languages \citep{vayani2025languagesmatterevaluatinglmms,wang2024not}), but given the multilingual capabilities of LLMs, it is important to put them to the test in a valid manner. For one thing, it can be informative to see how different scripts or more morphologically rich languages compare. This is of particular importance for low-resource languages, as they run the risk of being submerged in the digital ocean of English data. 

For low- and medium-resource languages, it is often hard to keep up with the pace and establish what already exists for high-resource languages. Benchmarking is no exception. As benchmark production with human input can be time-consuming and costly, one common approach in such settings is to machine-translate existing benchmarks for other languages or use LLMs to create synthetic benchmarks, for instance by generating question-answer pairs from pre-existing data such as Wikipedia. But what are the drawbacks to this approach and do they affect low-resource languages disproportionally?

In this paper we look at the current state of benchmarking for Icelandic as a case study of a low/medium-resource language, discuss the effects and results including machine-translated and synthetic data, and present the results of a quantitative error analysis on existing benchmarks for Icelandic. They show clear differences between human-authored or human-translated benchmarks vs. synthetic or machine-translated benchmarks, especially where the output has not been verified and/or corrected.\footnote{All code and data available at \url{https://github.com/finaing/who_benchmarks_the_benchmarks}}

\section{Related Work}
\label{sec:relatedwork}

Discussion on the pitfalls and problems of language model benchmarking, and the question of how to best compare the capabilities of different models, has been ongoing in the NLP community for years. At least as early as \citet{bowman-dahl-2021-will}, it had become apparent that the best-performing NLU models of the day, though far from perfect in real-world applications, had almost completely saturated existing benchmarks, begging the question of how best to measure actual progress. Four years later, this is still an open and important question as evidenced by \citet{eriksson2025ai} which identifies `key shortcomings' in benchmarking practices over the last decade, including `issues in the design and application', `misaligned incentives', `construct validity issues,' and other shortcomings directly related to the points made in this paper.

As previously discussed, it has become commonplace practice in recent years to automate the construction of benchmarks and evaluation sets either fully or partially. Here, statistical power (which is certainly important, see e.g. \citet{card-etal-2020-little}) is seemingly prioritized over precision and overview, despite the fact that machine learning models are frequently `right for the wrong reasons' \citep{mccoy-etal-2019-right}, capable of achieving high scores through learning unintended shallow heuristics. Indeed, there is considerable evidence that smaller, carefully curated test sets are often equally good or better indicators of performance than larger, synthetic sets (see e.g. \citet{gardner-etal-2020-evaluating,shaib-etal-2024-much,yadav-etal-2024-pythonsaga}). This is in line with the importance of \textit{validity} in any kind of testing. Present already in e.g. \citet{Cronbach1955}, validity is a key concept in psychometrics and addresses whether a test or evaluation instruments truly measures what it intends to measure (the construct). In the context of LLM testing and particularly in clinical applications, calls have been made to prioritize construct validity and employ a Benchmark-Validation-First Approach to (medical) LLM evaluation \citep{alaa2025}. To paraphrase, benchmarks need to be right for the right reasons.

Even automatically constructing test items using a structured template, now the norm for grammatical evaluation sets \citep{warstadt-etal-2020-blimp-benchmark,huebner-etal-2021-babyberta,song-etal-2022-sling,jumelet2025multiblimp10massivelymultilingual}, can easily result in semantically bizarre examples and items solvable through superficial heuristics rather than genuine linguistic competence \citep{javier-vazquez-martinez-etal-2023-evaluating}. Similarly, the widespread practice of evaluating systems for different languages by automatically translating existing benchmarks (usually from English) has repeatedly been shown to result in translation artifacts which affect accuracy and omit language- and culture-specific context \citep{10.1162/tacl_a_00317,park-etal-2024-translation,chen-etal-2024-good-data,liu-etal-2025-seaexam}. For instance, \citet{cengiz2025evaluatingqualitybenchmarkdatasets} examine 17 Turkish datasets (a “mid-resource” language), many of which are translations/adaptations from English. They identify major shortcomings related to coherence, fluency, cultural alignment, and increased risk of bias and error. Similarly, \citet{yao-etal-2024-benchmarking} note major shortcomings in culturally specific machine translations where exact translation pairs might not exist. Furthermore, \citet{semenov_sennrich_disfluency} demonstrate shortcomings of template-based translations in multilingual benchmarks, showing that sentence-level translations yield more reliable results.



As noted by \citet{vayani2025languagesmatterevaluatinglmms}, multilingual benchmarks remain limited in scope and disproportionately represent high-resource languages, neglecting the cultural and linguistic diversity of lower-resource languages. Moreover, benchmarks are often absorbed into training corpora, resulting in contamination and inflated evaluation scores \cite{yang2023rethinkingbenchmarkcontaminationlanguage}.

When evaluation datasets rely on translations of English benchmarks without cultural adaptation, models may optimize for unnatural constructions rather than authentic usage. This issue is amplified in small-language settings, where limited corpora increase the influence of benchmarks on model optimization. Indigenous languages in particular face severe digital resource scarcity, both in corpus size and representativeness. As \citet{wiechetek2024ethical} argue, when developers lack proficiency or active engagement with the target language, systematic errors may go unnoticed. Scarcity increases reliance on synthetic data, which can recursively dominate future training cycles, amplifying distortions and reinforcing inaccuracies. In extreme cases, synthetic outputs may even invert the meaning of source texts, causing direct harm to the language community.

Consequently, models may perform well on translated test sets while failing to capture culturally grounded semantics or authentic discourse patterns, leading to systematic overestimation of model capabilities \cite{sainz2023nlp}. This concern is supported by \citet{wang2024seaeval}, who show that LLM performance is inconsistent across languages, with degradation in lower-resource settings and stronger results on English cultural reasoning tasks. Similarly, \citet{wang2024not} demonstrate that LLMs tend to reproduce English-centric cultural assumptions even when prompted in other languages. Direct translation of culturally specific English benchmarks may therefore reward English-oriented reasoning rather than culturally situated understanding in the target language.

\section{Leaderboards for Icelandic}
\label{sec:motivation}
As of October 2025, there are two publicly available leaderboards for Icelandic. One is maintained by the Icelandic LT company, Miðeind, on Hugging Face\footnote{\url{huggingface.co/spaces/mideind/icelandic-llm-leaderboard}} and the other is a part of the EuroEval project, previously ScandEval \cite{nielsen2023scandeval}.\footnote{\url{euroeval.com}} 

\subsection{Miðeind's leaderboard}
At the time of writing, Miðeind's leaderboard evaluates 59 LLMs across 6 different benchmarks:

  \begin{itemize}
  \small
  \item \textit{WinoGrande-IS}: a human-translated and localized version of the WinoGrande test set \cite{snaebjarnarson-etal-2022-warm, 10.1145/3474381}
\item \textit{GED}: binary sentence-level GED with data from the Icelandic Error Corpus \cite{arnardottir-errorcorpus}
  \item \textit{Inflection}: inflection of 300 Icelandic adjective-noun pairs in all four cases and both numbers\footnote{\url{huggingface.co/datasets/mideind/icelandic-inflection-all-flat}}
  \item \textit{Belebele}: Icelandic subset of Belebele (reading comprehension) \cite{Bandarkar_2024}\footnote{\url{huggingface.co/datasets/facebook/belebele/viewer/isl_Latn}}
  \item \textit{ARC-Challenge-IS}: machine-translated version of the ARC-Challenge benchmark \cite{Clark2018ThinkYH}\footnote{\url{huggingface.co/datasets/mideind/icelandic-arc-challenge}} 
  \item \textit{WikiQA-IS}: 1.9k QA pairs created by GPT-4o from Icelandic Wikipedia data, manually verified and corrected \cite{thorunn-arnardottir-etal-2025-wikiqa}
    \end{itemize}

\subsection{EuroEval's leaderboard}
EuroEval's leaderboard contains benchmarks and evaluation for 18 European languages, at the time of writing, and evaluates all types of language models, fine-tuning encoder models and either zero- or few-shot prompting decoder models for different tasks.

Like Miðeind, EuroEval uses WinoGrande-IS and Belebele, but a different machine-translated version of the ARC-Challenge. It additionally uses WikiQA-IS in a different format, where an LLM is used to generate three plausible answer alternatives, turning it into a multiple-choice answering task, but the benchmark has not been published in this format. Other datasets or benchmarks included are:

\begin{itemize}
\small
  \item \textit{Hotter and Colder Sentiment}: sentiment analysis dataset \cite{fridriksdottir-etal-2025-hotter}
  \item \textit{MIM-GOLD-NER}: NER dataset \cite[paper]{ingolfsdottir-NER}, \citelanguageresource[data]{20.500.12537/140}
  \item \textit{ScaLA-IS} \cite{nielsen2023scandeval}, \textit{IceEC} (sampled data from the Icelandic Error Corpus), subset of \textit{IceLinguistic} \cite{armannsson-etal-2025-icelandic}: All used for a binary evaluation of grammaticality or GED
  \item \textit{NQiI}: ``Natural Questions in Icelandic'', dataset for open domain QA with retrieval
 \cite{snaebjarnarson-einarsson-2022-natural}\footnote{\url{huggingface.co/datasets/vesteinn/icelandic-qa-NQiI}}
  \item \textit{MultiWikiQA-IS}: created using LLM Gemini-1.5-pro on Wikipedia data (306 languages) \cite{smart2025multiwikiqareadingcomprehensionbenchmark}\footnote{\url{huggingface.co/datasets/alexandrainst/multi-wiki-qa}}
  \item \textit{MMLU-IS}: machine-translated version of the US MMLU benchmark \cite{hendrycks2021measuring}\footnote{\url{huggingface.co/datasets/alexandrainst/m_mmlu}}
  \item \textit{HellaSwag-IS}: machine-translated version of the HellaSwag benchmark \cite{zellers-etal-2019-hellaswag}\footnote{\url{huggingface.co/datasets/alexandrainst/m_hellaswag/viewer/is}}
  \item \textit{RRN}: summarization dataset \cite{sverrisson-einarsson-2023-abstractive}\footnote{\url{huggingface.co/datasets/thors/RRN}}
\end{itemize}

\noindent It is important to note two things. First, in some cases datasets are being turned into benchmarks, despite not being intended as such originally. This goes for \textit{Hotter and Colder}, \textit{MIM-GOLD-NER}, \textit{IceEC}, \textit{NQiI} and \textit{RRN}. Second, EuroEval labels some benchmarks as ``unofficial''. As far as we can see, there is no information to be found on the website on what factors are decisive in making this distinction between benchmarks, but the label apparently refers to benchmarks for which the results are not published on the leaderboard. ``Official'' benchmarks are: \textit{Hotter and Colder Sentiment}, \textit{MIM-GOLD-NER}, \textit{SCaLA-IS}, \textit{NQiI}, \textit{IcelandicKnowledge}, \textit{WinoGrande-IS} and \textit{RRN}. However, the published code on Github and the EuroEval Python package include scripts to create all of the benchmarks (and splits), regardless of whether they are unofficial or not, and benchmark models on them.\footnote{\url{github.com/EuroEval/EuroEval}}

\begin{table*}
\scriptsize
\centering
\begin{tabular}{|c|c|c|c|c|c|c|}
\hline
\textbf{Benchmark} & \textbf{Samples} & \textbf{Total size} & \textbf{Synth} & \textbf{IRR} & \textbf{CI} & \textbf{p-value} \\
\hline

ARC-IS (EuroEval)          & 112 & 1,116  & x  & 0.39& (0.28, 0.51) & $<.001$ \\ \hline
ScaLA-IS             & 102 & 1,024  &  & 0.52& (0.36, 0.67) & $<.001$  \\ \hline
IceEC                & 250 & 58,239 &  & 0.63& (0.55, 0.71) & $<.001$  \\ \hline
IcelandicQA          & 200 & 2,000  &  & 0.54& (0.39, 0.68) & $<.001$   \\ \hline
Belebele-IS          & 90  & 900   &  & 0.49& (0.36, 0.63) & $<.001$  \\ \hline
MultiWikiQA-IS       & 250 & 5,004  & x  & 0.71& (0.66, 0.77) & $<.001$   \\ \hline
ARC-challenge-IS (Miðeind) & 112 & 1,119  & x & 0.61 & (0.50, 0.72) & $<.001$   \\ \hline
GED                  & 20  & 200   &  & 0.74& (0.14, 1) & 0.018  \\ \hline
NQiI                 & 213 & 2,132  &  & 0.73 & (0.66, 0.79) & $<.001$   \\ \hline
MMLU-IS              & 28  & 277   & x & 0.31& (0.01, 0.62) & 0.044   \\ \hline
WinoGrande-IS        & 109 & 1,088  &  & 0.54& (0.36, 0.72) & $<.001$  \\ \hline
HellaSwag-IS         & 250 & 9,368  & x  & 0.22& (0.12, 0.32) & $<.001$   \\ \hline
\end{tabular}
\caption{Overview of benchmarks, sample size, total size, inter-rater reliability (IRR) scores along with their confidence intervals and p-values}
\label{tab:size-IRR}
\end{table*}

\section{Quantitative Error Analysis}
\label{QuantitativeAnalysis}
A review of benchmarks that are currently being used for Icelandic revealed various problems in some of the datasets and, to some extent, in the way they are being used. To keep this paper from consisting of seemingly cherry-picked examples, we conducted a quantitative error analysis of subsets of most of the benchmarks described in Section \ref{sec:motivation}. The review did not include \textit{RRN}, \textit{MIM-GOLD-NER}, \textit{IceLinguistic} and \textit{Hotter and Colder Sentiment}. The first two for technical reasons and the other two to be impartial in our judgments, as some of the authors participated in their creation and we acknowledge the drawback of that. The \textit{Inflection} benchmark should have been included, but simply was not at this stage, which is unfortunate.  Furthermore, we did not include \textit{IcelandicKnowledge}, EuroEval's version of the \textit{IcelandicQA} dataset, that has been turned into a multiple-choice question-answering task and three new answer options are created by an LLM. As mentioned, the benchmark has not been published in this format and we did not make use of EuroEval's Python package to create our own version.

To keep the sample sizes of the datasets within feasible boundaries, we used the following approach: If the dataset was longer than 250 rows, we took 250 randomly selected examples, else randomly selected 10\%. Where different splits were available, we used the train split, and for \textit{NQiI} we only extracted rows that had an answer, like EuroEval, as rows with no answer are only meant to serve as negative examples in model training.

The samples were annotated by three of the authors of this paper, all of whom have a background in both Icelandic linguistics and NLP, and we used the following three labels and guidelines:

\begin{itemize}
\small
    \item IC: one of the following: incorrect (e.g. wrong answer to a question), severely flawed, clearly/poorly machine-translated  
    \item F: flawed example, contains errors (spelling, grammaticality, vocabulary)
    \item OK: Valid example
\end{itemize}

\noindent In the case of the question-answering benchmarks, \textit{IcelandicQA}, \textit{NQiI} and \textit{MultiWikiQA-IS}, we focused on the QA pairs in our error annotation and only considered the context to verify validity, i.e. the context itself is not evaluated. We admit that our labels could have been more fine-grained with regards to error types to perhaps give some kind of estimate of their effect on the evaluation. Although the IC-label was intended to cover all examples that would potentially affect the evaluation validity there is a clear difference between factually wrong answers and very poor machine translations. The former will certainly affect the evaluation whereas an LLM might still be able to ``make sense'' of the latter and provide the correct answer.

An overview of the sample size, original length and the inter-rater reliability (IRR) with confidence intervals is shown in Table \ref{tab:size-IRR}, where the x-label in the column `synth' indicates whether the benchmark was either machine-translated or created by an LLM. Note that IcelandicQA does not have such a label as the model output was manually verified and corrected where necessary. For calculating the IRR, we use Krippendorff's alpha and weighted agreement coefficients (ordinal) to account for the ordinal nature of the ratings, i.e. IC > F > OK.

For each of the benchmarks we have three annotations and overall the level of agreement is \textit{good} (following \citet{Regier2013}), \textit{questionable} for ARC-IS (EE), MMLU-IS and HellaSwag-is, and \textit{very good} for MultiWikiQA-is, ARC-Challenge-is, GED and NQiI. We note that the low IRR score in HellaSwag-IS is only due to disagreement between the IC and F labels as the annotated sample has no valid examples. This has, however, not been analyzed any further and the IRR values should be taken into account for our interpretation of the results. 

\subsection{Results}
\label{subsec:results}
\begin{figure*}[t]
    \centering
    \includegraphics[width=1\linewidth, trim={0 2cm 0 0},clip]{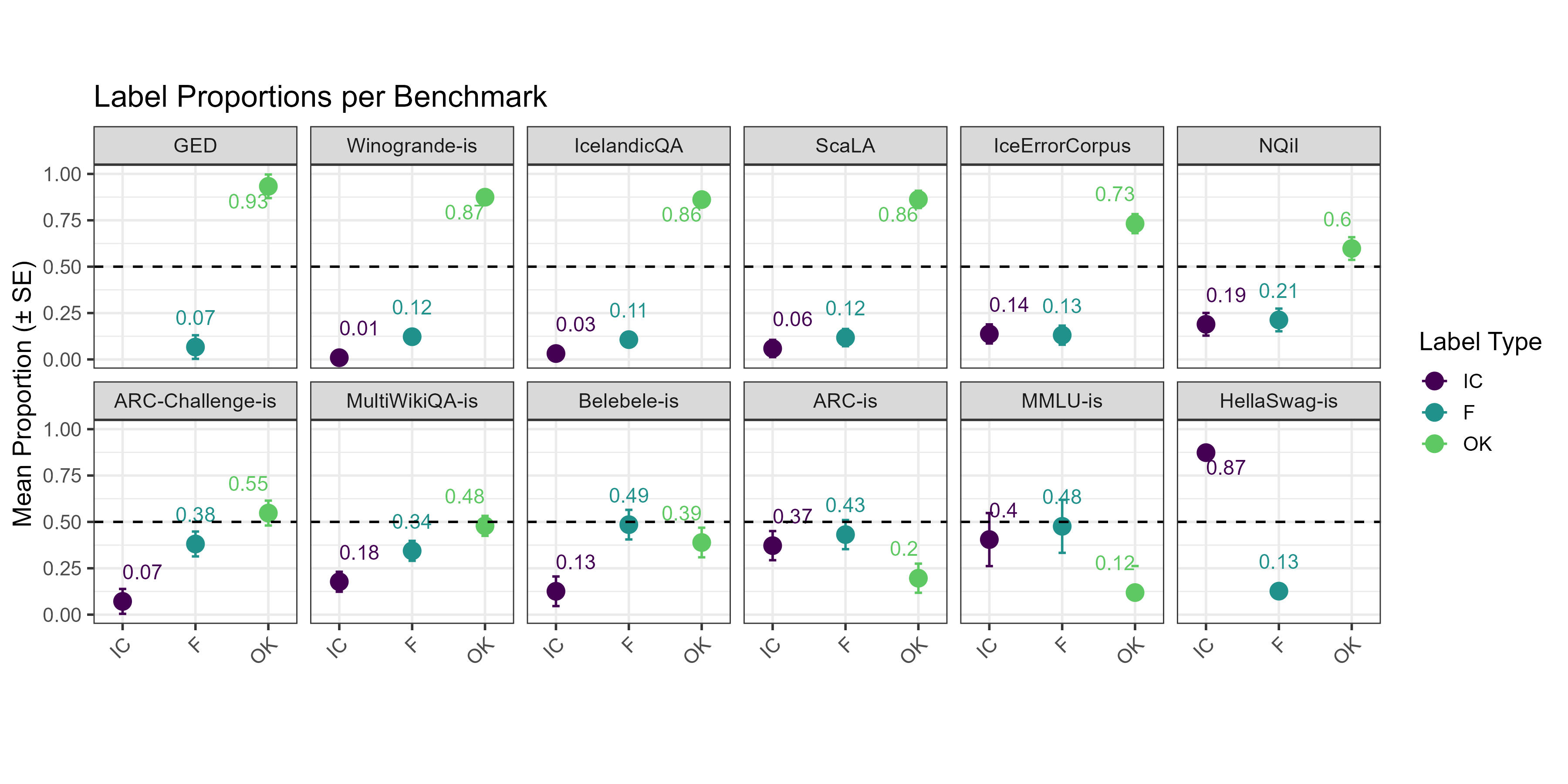}
    \caption{Mean proportion and 95\% confidence interval for each of the labels across annotations and benchmarks, arranged by OK proportion in descending order}
    \label{fig:distribution}
\end{figure*}

The results of our error annotation are shown in Figure \ref{fig:distribution}, where the mean proportion of each label across all raters is shown for each benchmark with a 95\% confidence interval. In Appendix \ref{appendix:label_props} we show the label proportions for each rater across the benchmarks.

First, we observe a clear difference in benchmarks according to whether or not humans have been involved in their creation, translation or verification. Based on our samples, we find \textit{Hellaswag} to be almost entirely severely flawed and \textit{MMLU} very flawed, with less than 10\% of the sample being valid (admittedly, the sample size for \textit{MMLU-IS} could have been bigger and the inter-rater agreement is questionable based on our interpretation). Furthermore, we find that the machine-translated version of ARC that EuroEval uses (ARC-is) has an OK proportion of only 20\%, whereas the localized version Miðeind uses (ARC-Challenge-is) is better, with a much lower IC-proportion (7\% compared to 37\%) and a higher OK-proportion (55\%).

For \textit{MultiWikiQA-IS}, less than half of the examples in our sample receive the OK-label. Our error labels admittedly make no distinction between factual and linguistic errors, but both types of flaws can affect the benchmark's validity. In future evaluations, we aim to distinguish between these two types of flaws although they can sometime appear in conjunction.  The following example illustrates the type of problematic items in \textit{MultiWikiQA-IS}: An article about the Icelandic philosopher Salvör Nordal is presented as context and the question is (in our translation): ``Which married couple had the son Sigurjón Nordal?'' The male name, Sigurjón, does not appear in the article and the only Icelander to ever have had that name is completely irrelevant to the context, but the answer to the question is Salvör's parents. In contrast, IcelandicQA, where QA pairs generated by an LLM were manually verified and corrected, is much better, although it contains validity errors as well.

Surprisingly, the Icelandic version of Belebele has a rather low proportion of OK examples. In the paper by \citet{Bandarkar_2024}, the authors state that it ``was created end-to-end without the use of machine translation technology, relying solely on experts fluent in English and the target language.'' We have, however, reasons to believe that the translators involved in translating the data into Icelandic used MT services, at least to some extent. One blatant example is an item where one of the answer options, which only include geopolitical entities, is the Icelandic word \textit{kalkúnn} which translates to the animal turkey, not the country, in English. Here, \textit{Tyrkland} would have been the correct translation and this answer option also happens to be the correct one. No fluent expert in English and Icelandic would make such a mistake. A good contrast to these machine-translated benchmarks is the Icelandic version of the WinoGrande benchmark \cite{snaebjarnarson-etal-2022-warm}, which was translated by humans and localized. Based on our sample, it contains no severely flawed examples.

Regarding the results for NQiI \cite{snaebjarnarson-einarsson-2022-natural}, it is important to keep in mind that this is a case of a dataset being turned into a benchmark despite this not being the intended use -- it was originally created for open domain QA retrieval. As it is being used as a benchmark, however, we evaluate it as such. Based on our sample, there unfortunately seem to be some validity errors in the QA pairs. In one example, context about Kim Jong-Un is paired with a question about the capital of Bolivia, whereas the answer is related to the Kim Jong-Un (nqii\_3570). In another example, a badly-construed pair consists of a strange question about when tap water in Iceland was first boiled and the answer is that `it boils at 100°C' (nqii\_7841), where the context is an article about water in general (we refer to Section \ref{discussion} for more discussion).


\section{Discussion}
\label{discussion}
We would like to begin by stressing that the goal is not to create flawless benchmarks, but to improve current practices, both by revising and re-releasing existing benchmarks where possible and by designing stronger future ones. In light of our findings, we argue that it is important to change the way benchmarking is approached in lower-resource settings and we will start by addressing the use of MT.



\subsection{Machine-translated benchmarks}
In general, we find it questionable to use machine-translated versions of foreign/non-localized benchmarks, especially in low/medium-resource settings and even more so if the output is not verified or corrected. Such an approach immediately raises several questions: 1) How was the MT model chosen? 2) Is the quality of the output evaluated in any way? 3) How suitable is the original benchmark for benchmarking another language?

To start with the first question, one point of reference could be findings of WMT, the largest conference in the MT community, where Icelandic has been a part of the general shared MT task in the last two years. In 2024 (\citeauthor{kocmi-etal-2024-findings}), human translators achieved the highest score for for EN$\rightarrow$IS with 93.1 on a 0--100 human evaluation scale (with 0 being `no meaning preserved' and 100 being that the meaning and grammar is completely consistent with the source). The best performing ML system, Dubformer (a private translation service), scored 84.9, followed by Claude 3.5 Sonnet at 81.9. Incidentally, two different translations by Claude 3.5 Sonnet of the ARC-Challenge are being used for Icelandic evaluation by Miðeind and EuroEval. This raises the question of whether those translations can, out-of-the-box, be expected to be of better quality than 81.9\%, and whether that level is sufficient for benchmark construction.

As for the second question, the only MT quality assessment we identified for these benchmarks is the one conducted by \citet{smart2025multiwikiqareadingcomprehensionbenchmark} for MultiWikiQA. In that study, crowd workers rated the fluency of a small sample of translated items on a 1--3 scale, with Icelandic (n=150) scoring $\sim$2.68/3. Note that the fluency was meant to be evaluated \textit{regardless} of whether the question was ``unanswerable or require[d] context to be answered''. We refer to Section \ref{QuantitativeAnalysis} for our evaluation, but at this point we would like to note that fluency evaluation alone, without any measure of adequacy or accuracy, is unlikely to give a comprehensive assessment of the dataset.

\begin{table*}[ht!]
\fontsize{8pt}{8pt}\selectfont
\centering
\begin{tabular}{|p{0.5cm}|p{7cm}|p{7cm}|}
\hline
\multicolumn{1}{|c|}{\textbf{Ref}} &
  \multicolumn{1}{c|}{\textbf{Original}} &
  \multicolumn{1}{c|}{\textbf{MT / \textit{Our translation}}}
   \\ \hline
1 & Louis Pasteur created a process that reduced the amount of bacteria in milk. How does this process most likely benefit people? & \textbf{M:} Louis Guðmundsson fann upp aðferð sem dró úr fjölda baktería í mjólk. Hvernig nýtist þessi aðferð fólki líklega best?
\\ \hline
2a & Which protist is a bi-flagellate autotroph? & \textbf{M}: Hvaða frumvera er tvíflagella sjálfnærandi lífvera? \\ \hline
2b & - & \textbf{EE}: Hvaða mótefni er bi-flagellate autotroph? \\ \hline
3 & Hvers vegna erum við ekki með lögin þannig að hver sem opinberlega hæðist að, rógber, smánar og ógnar manni og svo framvegis án tillits \textbf{Forseti hringir.} til nokkurs skapaðar hlutar sæti refsingu? & \textit{Why are our laws not written such that anyone who publicly mocks, defames, humiliates and threatens someone etc., with no regard to \textbf{Speaker rings.} anything at all, is punished?}\\ \hline
4 & Varðandi smábátahlutfallið í þessu \textbf{frumvarpi} verð ég að segja ég hef ekki skoðað skiptinguna á því mjög „grundigt“. & \textit{Regarding the proportion of small fishing boats in this \textbf{bill}, I have to say that I have not looked into the distribution very thoroughly.} \\ \hline
\end{tabular}
\caption{Examples of flaws in Icelandic benchmarks. For space reasons we use the same column for either machine-translated examples from the benchmarks or, in italics, our translation of Icelandic examples for the foreign reader. (\textbf{M}: Miðeind, \textbf{EE}: EuroEval.)}
\label{tab:examples}
\end{table*}

Regarding the third question, we use the ARC-Challenge as an example. The original benchmark consists of human-authored, grade-school science questions from the US. One might ask how pertinent it is to evaluate LLM performance in Icelandic on it in the first place, regardless of whether it is human- or machine-translated, just as it would be questionable to make Icelandic grade-school pupils answer the same questions without accounting for cultural differences and some localization of content. To take a specific example from the original benchmark, in item Mercury\_7084648 the question is about the causes of low precipitation in Nevada. It is unclear how culturally or educationally relevant such a question is in Icelandic context, highlighting the broader issue of contextual misalignment in translated benchmarks.

As far as localization is concerned, in Miðeind's version, the MT model seems to have translated most names that occur in the benchmark -- presumably this was included in instructions to an LLM. At a quick glance, this seems to be a valid localization method, but it has some unintended side-effects as example 1 in Table \ref{tab:examples} shows, where the name Louis Pasteur is erroneously changed to \textit{Louis Guðmundsson}, removing all relation to the real person, which is of relevance for the question.

Another example of the shortcomings of this approach are items 2a and 2b in Table \ref{tab:examples}. In Miðeind's version, \textit{bi-flagellate} is translated as \textit{tvíflagella}, which is not an Icelandic word (\textit{tvísvipa} is correct), and \textit{autotroph} should be \textit{frumbjarga lífvera}, rendering the translation incorrect. EuroEval's version is only half-translated and the Icelandic translation of \textit{protist} is wrong. These examples show how the MT can fail on scientific terminology, which is unfortunate for a benchmark based on science questions. In light of this, can it be said that performance on such data meaningfully reflects a model's Icelandic reading comprehension abilities, if that is what the benchmark is intended to measure? Furthermore, it is concerning that the output has seemingly not been filtered prior to release. In EuroEval's version, one question (\href{https://huggingface.co/datasets/alexandrainst/m_arc/viewer/is/train?q=ARC-Challenge%2Ftrain%2FMercury_401011&views%5B%5D=is_train&sql=SELECT+*+FROM+is_train+WHERE+ID%3D%27ARC-Challenge%2Ftrain%2FMercury_401011%27%3B}{ID Mercury\_401011}) includes the Icelandic word \textit{gös} `gases' repeated 399 times, nothing else.\footnote{In the EuroEval code there are automatic checks in place to not include ``overly repetitive samples'' (and samples with overly short or long texts) but we are referring to the published data.}

Another example of a machine-translated benchmark is \textit{HellaSwag-IS}, excluded by Miðeind and marked ``unofficial'' by EuroEval. Even a cursory inspection reveals to any native speaker how poor the quality of translation is. Our quantitative error analysis (Section \ref{QuantitativeAnalysis}) confirms that the Icelandic version is severely flawed, and we argue that it should be discarded. The IRR score is admittedly low, but not a single item was labeled as valid by any of the annotators, which is telling for its quality.

\subsection{No native speaker involvement}
Another concern arises when benchmarks are developed without the involvement of native speakers of the target language. One such example is the ScaLA benchmark \cite{nielsen2023scandeval}, which covers the Nordic languages as well as German, Dutch and English. ScaLA measures linguistic acceptability through binary judgments of sentences that are either in their original form (``correct'' sentences) or artificially corrupted, either by deleting a word or by swapping two adjacent words. The data is from the Universal Dependencies framework\footnote{\url{https://universaldependencies.org}} and the authors assume the original sentences are grammatically correct. Although they describe steps taken to ensure that the corrupted sentences are indeed ungrammatical, this verification was conducted only on a random sample of Danish data.

The problems with the Icelandic subset of this benchmark are twofold. The former is related to peculiarities in the data, of which the creators of ScaLA were probably unaware. Most of the data consists of transcribed speeches from the Icelandic parliament. While the speeches are proofread and produced in a formal setting, it is still a word-for-word transcription of spoken language and not written text. While not necessarily ungrammatical, they often include sentences that are perhaps not ideal for such a test.

Moreover, the data includes procedural insertions such as a) \textit{Forseti hringir.} `Speaker rings' and b) \textit{Gripið fram í:} `Interruption', which mark, respectively, a) the moment when the speaker of the house rings a bell to indicate that the allocated time to an MP is up, and b) when someone interrupts a speech \textit{ex auditorio}. An example of a) is shown in 3 in Table \ref{tab:examples}. Although the sentence is labeled as correct, such insertions disrupt the syntactic structure of the surrounding speech. There are unfortunately also two instances in the benchmarks where these two phrases (a and b) are themselves subjected to artificial corruption through word-order reversal (\textit{hringir Forseti.} and \textit{fram Gripið í:}), which makes little sense as the phrases themselves are already corruptions of the sentence.


The second problem is that in some cases the Icelandic sentence is still grammatically correct after corruption. One such example is number 4 in Table \ref{tab:examples}. In the corrupted version, the noun in boldface is deleted, but the resulting sentence is just as correct and grammatical as the original. Lastly, there are examples of original sentences being grammatically wrong, resulting in erroneous labels. 

A related issue concerns EuroEval's use of the Icelandic Error Corpus \citelanguageresource{20.500.12537/105} for binary grammatical error detection (GED). The corpus consists of proofread texts from various sources, with corrections made independently by different proofreaders, unfortunately resulting in some inconsistencies. Errors were classified in five different categories: \textit{coherence}, \textit{grammar}, \textit{orthography}, \textit{style}, \textit{vocabulary}. EuroEval, however, samples the data without regard to error type and treats the `has\_error' label as an indicator of the grammaticality of a sentence. This is problematic. A stylistic correction -- of which there 39 examples in our sample -- hardly changes an ungrammatical sentence to a grammatical one or vice versa. For GED purposes, filtering the data by relevant categories such as grammar or orthography would be more appropriate, which Miðeind seems to have done in their use of the same data. Similarly, there are instances where there is an error in a sentence that is supposed to be error-free, which again leads to erroneous labels, as well as errors in corrected sentences, but the latter are not used by EuroEval so they have no bearing on the benchmarking.

These deficiencies in both benchmarks do not mean that they are ``bad'', but simply show how their creation or use as benchmarks could have been improved with native speaker involvement and proper validation of the data for these purposes. We would like to add that in both cases, the data that makes up these benchmarks was collected or created by native speakers but it is their use as benchmarks that is under scrutiny. 

\subsection{MT without native speaker involvement}
Although it is not a part of public leaderboards for benchmarking in Icelandic, a recent benchmark illustrates similar concerns, namely \textit{GAMBIT+: A Challenge Set for Evaluating Gender Bias in Machine Translation Quality Estimation Metrics} by \citet{filandrianos-etal-2025-gambit}. There, Icelandic is used as one example of a grammatically gendered language (along with 10 others). For each English source text with gender-ambiguous occupation terms, two target language translations are created, intended to differ ``only in the grammatical gender of the occupation and the dependent grammatical elements''. If, for instance, the English source sentence includes the word \textit{professor}, the two target sentences should differ only in the gender of the translation of that word (and its dependent elements). The assumption is therefore that this is possible in the languages involved, but that is simply not the case for Icelandic, as such minimal gender pairs in occupation terms are few (see e.g. \citet{fridriksdottir-einarsson-2024-gendered} where 381 out of 394 occupation terms examined were masculine, eight feminine and five neuter). The word \textit{professor} is used in the example taken in the paper to demonstrate the approach for all the languages included. In Italian they use \textit{il professore} : \textit{la professoressa}, in Serbian \textit{profesor : profesorica}, etc. But in the Icelandic example, the sole difference lies in the use of the titles \textit{Sir} (icel. \textit{Herra}) and \textit{Mrs.} (icel. \textit{Frú}) with the noun (``Herra/Frú prófessorinn''), as no feminine version of the word exists. These titles are, however, very rarely used in modern Icelandic and of course do not constitute a modification of the grammatical gender of the target noun. 

In Icelandic, masculine occupation nouns are generally used irrespective of the referent's gender, though some feminine counterparts exist in compounds, such as \textit{forstöðumaður : forstöðukona} `director man/woman (of e.g. an institute)'. In the dataset, however, many minimal pairs rely on wrong or hallucinated feminine forms. Examples of this include \textit{rithöfundarins : rithöfundarkonunnar}, hallucinated feminine term for \textit{writer}; \textit{puntari : puntara}, hallucinated translation of \textit{prompter} (in a theater) regardless of gender; and \textit{kjarnorkuverskjörinn} : \textit{kjarnorkuverskjörinn (fem.)}, where the translation of the occupation itself (\textit{nuclear power plant operator}) is hallucinated, the words are identical, and the feminine gender is added in parenthesis (\textit{fem.}) to indicate a gender difference that does not exist.

Third, it is important to note that masculine agreement in pronouns, adjectives etc. with masculine nouns is grammatically correct in Icelandic regardless of the referent's gender. Feminine agreement may, however, be used to specifically express the feminine gender of the person (and similarly, the neuter, for a non-binary person). There are some Icelandic examples in the benchmark of this form, i.e. where the only difference lies in agreement with the noun but not the noun itself. However, as the authors say, the target translations were ``constructed to be semantically and lexically identical, differing only in the gender marking of the occupation and the dependent grammatical elements.'' It thus seems that examples with agreement only in dependent grammatical elements and not in the gender marking of the occupation do not align with the authors' intention, further illustrating how ill-suited Icelandic is for this approach.

The translations were generated with Claude Sonnet 3.5 and verified with Claude Sonnet 3.7 (same family as the translation model) to ensure that the gender of the occupation was the only difference between the two target sentences and Icelandic achieved the lowest accuracy among the included languages (86.75\%) in this validation. On the face of the arguments presented here, we believe that the quality of the Icelandic data is actually lower than this percentage indicates. A more rigorous and quantitative evaluation of the benchmark is needed to verify this, but it is already clear that the methodology can lead to serious flaws in the benchmark. We would lastly like to note that this dataset is intended to evaluate gender bias in MT and the authors claim to reveal ``statistically significant score shifts driven solely by grammatical gender [...] with a clear overall tendency to favor masculine forms''. As far as Icelandic is concerned, we wonder to which extent these results are themselves biased by enforced and hallucinated feminine forms as in the examples above.


\subsection{Typos, errors and whether they matter}

LLMs are robust to linguistic errors, so why be finicky about that? The answer depends on the purpose and nature of the benchmark. Consider the NQiI dataset, where \textit{natural questions} are meant to indicate naturally occurring questions, such as a user might ask an LLM or search engines. In practice, such questions can and do often contain errors that do not necessarily prevent the system from correctly interpreting or answering the question. This can also be useful as training data for models, to expose them to a wider variety of language than only the standard. However, our focus is not on training utility but on the use of such datasets as benchmarks. In cases like this, with QA benchmarks, we are more concerned with validity errors, as in the examples in Section \ref{subsec:results}, and not natural linguistic variation by humans. On the other hand, the machine-translated versions of ARC originate from carefully constructed US grade-school science questions, presumably well-formed and appropriate for the task in English. This raises a different concern, whether it is acceptable to evaluate models on translated versions that introduce linguistic distortions through ``translationese'' or outright errors which no native speaker would accept.

A broader question, which we leave for future work, is to what extent ``translationese'', LLM hallucinations or flaws in benchmarks matter. Someone might contend that the most of the benchmarks involved in this paper, regardless of their flaws, are sufficient to show differences in model performance. But to what extent do the flaws affect the outcome? And does it matter if they do? As an example, we can think of a benchmark that has 80\% valid examples and 20\% severely flawed. Given a large enough sample size for sufficient statistical power, are the 20\% ``acceptable'' if the rest shows clear difference in model performance? We argue that, regardless of short-term performance comparisons, flawed benchmarks can be harmful if incorporated into optimization pipelines, thereby influencing model development (see Section \ref{sec:relatedwork}). In general, we believe the goal should always be to design benchmarks which are right for the right reasons.

With increased use of automated methods, ever larger benchmarks can be built and with increasing size, it becomes increasingly unlikely that any human will ever read through all of the examples for verification. The resulting loss of oversight makes systematic quality control even more essential, even if it is only via stratified random sampling.

\subsection{Beyond single-label output, towards more diverse benchmarks}

As mentioned in Section \ref{sec:intro}, there is demand for creating new benchmarks that are better adapted to the capabilities of LLMs and go beyond evaluating single-label output. Most of the benchmarks involved in our evaluation are indeed such benchmarks, where the answer is either True/False or an answer option in a multiple-choice task, with only the QA benchmarks as an exception. 

We also note that, overall, the benchmarks included in our evaluation are quite similar. \textit{GED}, \textit{ScaLA} and \textit{IceErrorCorpus} all measure very similar phenomena with binary outcomes, raising questions about the added value of reporting them separately. Similarly, the three QA benchmarks are all Wikipedia-based, although there is not necessarily overlap between them, and MultiWikiQA-IS was created despite the existence of \textit{IcelandicQA}, that was created in a similar way but manually verified and corrected, which MultiWikiQA was not. There are two different versions of ARC, which are reading comprehension benchmarks just like \textit{Belebele} and \textit{MMLU}. \textit{WinoGrande-IS} targets commonsense reasoning, while \textit{HellaSwag-IS}, as we argued above, should likely be discarded. It would therefore be of benefit to try and broaden the spectrum of benchmarks for Icelandic and include more diverse tests, for instance with respect to the text generation capabilities of LLMs. This is to some extent done in three of the benchmarks that were not included in our evaluation, \textit{Inflection}, the \textit{Icelandic Linguistic Benchmark} and \textit{RRN} (see Section \ref{sec:motivation}).


\section{Conclusion}

In light of our findings, we would like to summarize a few key points to which attention should be paid in our opinion when benchmarking low/medium-resource languages: 

\begin{itemize}
  \item Avoid use of machine-translated benchmarks, especially when MT quality is still not sufficiently fluent, and/or when the output (or at least a sample of it) has not been verified and corrected
  \item Using an LLM to generate data can be tantamount to machine-translating data in terms of text quality and the same caveats apply
  \item Involve native speakers and prioritize the use of native-authored data
\end{itemize}

\noindent As already outlined, our results show clear differences with regards to native speaker involvement compared to machine-translated data where the output has not been verified or corrected, and it is clear that the involvement of native speakers and native data makes a difference for the better. In the Icelandic context, further opportunities exist to develop native benchmarks, for example by drawing on human-authored and validated materials from the school system (\textit{à la} the ARC Challenge and many more). 

As some of the authors of this paper have themselves been involved in the creation of benchmarks for Icelandic, we acknowledge that building good benchmarks is a complicated task and that we are not exempt from criticism of the problems outlined in the overall context of this paper. It is our hope that this can be a contribution to a healthy debate on how to proceed and what to avoid. We emphasize the need for native-authored, community-vetted, and contamination-audited benchmarks, ideally developed through participatory design with native speakers. This approach ensures that evaluation metrics genuinely reflect linguistic and cultural specificities of the language, preventing models from optimizing for superficial patterns introduced by translation artifacts or data leakage.


Lastly, we return to the question posed in the title of this paper: \textit{who benchmarks the benchmarks?} The responsibility must lie with their creators. Benchmark construction cannot be treated as a neutral or one-time contribution, it requires continuous scrutiny, documentation, and critical evaluation. The work presented here originated from a simple review of existing Icelandic datasets, where repeated irregularities and inconsistencies prompted the systematic analysis reported in this paper. Such examination should not be incidental, it should be standard practice.

\section{Limitations}
We first recognize that the annotation may have been biased by the overall topic of this paper, i.e. flaws in benchmarking, as well as for benchmarks where the annotators knew that they were machine-translated or synthetic. Secondly, the sample size could have been bigger, especially for the 10\% samples of benchmarks. Third, in three cases, the level of agreement was only questionable which must be taken into account. Fourth, the interpretation of the border between the error labels `IC' (severely flawed) and `F' (flawed) may have varied from one annotator to the other. Furthermore, the labels do not necessarily reflect whether the error is on the factual or linguistic level (although `IC' was meant for examples that are flawed to such an extent that it potentially affects the results validity). Moreover, as mentioned in Section \ref{QuantitativeAnalysis}, two benchmarks that are included in leaderboards for Icelandic were left out in the evaluation for impartiality reasons. We concede that this is a drawback and they should preferably be evaluated as well. Finally, our study evaluates benchmark quality rather than directly quantifying the downstream impact of these flaws on model rankings or comparative conclusions. Future work should investigate to what extent the identified issues materially alter evaluation outcomes.

\section{Acknowledgements}
We would like to thank our reviewers for useful comments and feedback as well as those who gave their opinion on the paper in earlier versions. This project was in part funded by the Language Technology Programme for Icelandic 2024–-2026. The programme, which is managed and coordinated by Almannarómur, is funded by the Icelandic Ministry of Education, Science and Culture. Steinunn Rut Friðriksdóttir was supported by the Eimskip Fund of The University of Iceland no. HEI2025-96428 and the Ludvig Storr Trust no. LSTORR2023-93030.

\section{Bibliographical References}\label{sec:reference}

\bibliographystyle{lrec2026-natbib}
\bibliography{heimildir}

\begin{thebibliography}{2}
\expandafter\ifx\csname natexlab\endcsname\relax\def\natexlab#1{#1}\fi

\bibitem[{Ingason et~al.(2021)Ingason, Stef{\'a}nsd{\'o}ttir, Arnard{\'o}ttir, and Xu}]{20.500.12537/105}
Anton~Karl Ingason, Lilja~Bj{\"o}rk Stef{\'a}nsd{\'o}ttir, {\TH}{\'o}runn Arnard{\'o}ttir, and Xindan Xu. 2021.
\newblock \href {http://hdl.handle.net/20.500.12537/105} {Icelandic error corpus ({IceEC}) version 1.1}.
\newblock {CLARIN}-{IS}.

\bibitem[{Ing{\'o}lfsd{\'o}ttir et~al.(2020)Ing{\'o}lfsd{\'o}ttir, Gu{\dh}j{\'o}nsson, and Loftsson}]{20.500.12537/140}
Svanhv{\'{\i}}t~Lilja Ing{\'o}lfsd{\'o}ttir, {\'A}smundur~Alma Gu{\dh}j{\'o}nsson, and Hrafn Loftsson. 2020.
\newblock \href {http://hdl.handle.net/20.500.12537/140} {{MIM}-{GOLD}-{NER} – named entity recognition corpus (21.09)}.
\newblock {CLARIN}-{IS}.

\end{thebibliography}

\section{Language Resource References}
\label{lr:ref}
\bibliographystylelanguageresource{lrec2026-natbib}
\bibliographylanguageresource{languageresource}

\appendix
\onecolumn
\section{Appendix: Label Proportions per Rater}
\label{appendix:label_props}
\begin{figure*}[h]
    \centering
    \makebox[\linewidth]{
        \includegraphics[width=1.5\linewidth]{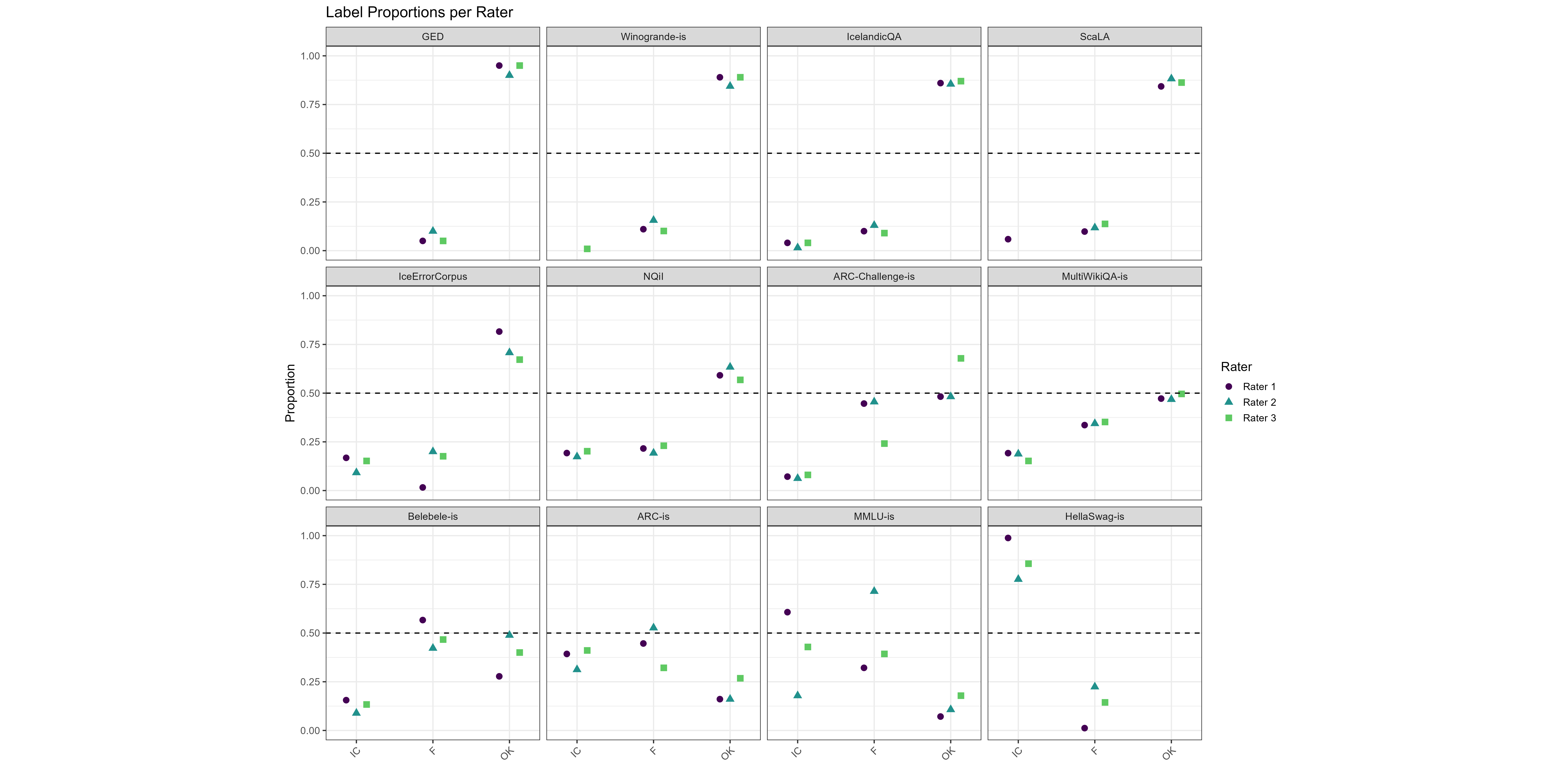}
        }
    \caption{Label proportions per rater across the evaluated benchmarks}
    \label{fig:props_per_rater}
\end{figure*}

In Figure \ref{fig:props_per_rater} we show the distribution of labels per rater across the benchmarks.


\end{document}